\documentclass{article}

\usepackage{graphicx}    
\usepackage{subcaption}  
\usepackage{microtype}
\usepackage{graphicx}
\usepackage{booktabs} 

\usepackage{hyperref}

\usepackage[ruled,vlined,algo2e]{algorithm2e}

\usepackage[accepted]{icml2025}

\usepackage{amsmath}
\usepackage{amssymb}
\usepackage{mathtools}
\usepackage{amsthm}
\usepackage{bm}

\theoremstyle{plain}
\newtheorem{theorem}{Theorem}[section]

\newtheorem{lemma}[theorem]{Lemma}

\theoremstyle{definition}
\newtheorem{definition}[theorem]{Definition}

\theoremstyle{remark}

\usepackage[textsize=tiny]{todonotes}

\icmltitlerunning{FedSTaS: Client Stratification and Client Level Sampling}

\begin{document}

\twocolumn[
\icmltitle{FedSTaS: Client Stratification and Client Level Sampling  \\
           for Efficient Federated Learning}



\icmlsetsymbol{equal}{*}

\begin{icmlauthorlist}
\icmlauthor{Jordan Slessor}{uofa}
\icmlauthor{Dezheng Kong}{uofa}
\icmlauthor{Xiaofen Tang}{uofa}
\icmlauthor{Zheng En Than}{uofa}
\icmlauthor{Linglong Kong}{uofa}
\end{icmlauthorlist}

\icmlaffiliation{uofa}{Department of Mathematical and Statistical Sciences, University of Alberta, Edmonton, Alberta, Canad}

\icmlcorrespondingauthor{Jordan Slessor}{jaslesso@ualberta.ca}

\icmlkeywords{Machine Learning, ICML}

\vskip 0.3in
]



\printAffiliationsAndNotice{\icmlEqualContribution} 

\begin{abstract}
Federated learning (FL) is a machine learning methodology that involves the collaborative training of a global model across multiple decentralized clients in a privacy-preserving way. Several FL methods are introduced to tackle communication inefficiencies but do not address how to sample participating clients in each round effectively and in a privacy-preserving manner. In this paper, we propose \textit{FedSTaS}, a client and data-level sampling method inspired by \textit{FedSTS} and \textit{FedSampling}. In each federated learning round, \textit{FedSTaS} stratifies clients based on their compressed gradients, re-allocate the number of clients to sample using an optimal Neyman allocation, and sample local data from each participating clients using a data uniform sampling strategy. Experiments on three datasets show that \textit{FedSTaS} can achieve higher accuracy scores than those of \textit{FedSTS} within a fixed number of training rounds.
\end{abstract}

\section{Introduction}
Federated learning is a methodology that allows clients to collaboratively learn a global model while preserving the privacy of their local data. At a high level, FL operates as follows: a central server sends an estimate of the global parameter to participating clients, who compute updates based on this parameter and their data. The clients then return the updated parameters to the server, which are aggregated to compute a new global parameter. This iterative process continues until the model converges. Several challenges accompany FL, including communication bottlenecks, system and client heterogeneity, and privacy concerns.

One approach to addressing both communication bottlenecks and client heterogeneity is \textit{FedAvg} \cite{fedAvg}. \textit{FedAvg} involves sampling a subset of clients in each training round and allowing these clients to perform multiple local training iterations before sending their model updates to the server. Despite its communication advantages, \textit{FedAvg} introduces bias into the global model. To address this issue, \textit{FedProx} \cite{FedProx}, a generalization of \textit{FedAvg}, was proposed. \textit{FedProx} incorporates a proximal term which limits the deviation of local updates from the global model, addressing the bias. Additionally, \textit{FedProx} provides convergence guarantees for training under non-IID data, a common scenario in FL.

While methods like \textit{FedAvg} and \textit{FedProx} address critical challenges in communication and bias, how to effectively sample the participating clients in each round remains an open problem. To illustrate the need for good client level sampling, consider having each client participate in training. In doing so, there's a reasonable likelihood for similar data being use repetitively, leading to redundant training. However, if the set of participating clients is sampled well, we may avoid some of these redundancies. Clustering-based approaches, including \cite{FedClust}, \cite{FedStrat}, and \cite{FedSTS}, focus on improving the efficiency of client sampling by grouping clients with similar characteristics. In \textit{FedSTS} \cite{FedSTS}, clients are grouped based on a compressed representation of their gradients and sampled using stratified random sampling techniques. Moreover, the number of clients sampled from each stratum is re-allocated using an optimal Neyman allocation, which minimizes the variance introduced by client sampling and significantly improves convergence rates. Data-level sampling methods, such as \textit{FedSampling} \cite{FedSampling}, aim to mimic centralized learning by sampling local data from each participating client, followed by only using this data in the local training.

In this paper, we propose \textbf{F}ederated \textbf{S}tratification \textbf{a}nd \textbf{S}ampling (\textit{FedSTaS}), a client and data-level sampling approach inspired by \textit{FedSampling} and \textit{FedSTS}. A key contribution of \textit{FedSTaS} is the integration of compressed gradient clustering from \textit{FedSTS} with the privacy-preserving data-level sampling methodology of \textit{FedSampling}. This allows \textit{FedSTaS} to achieve three key benefits: reduced variance, improved training efficiency, and privacy protection through the use of local differential privacy (LDP). Theoretically, we show the unbiasedness of \textit{FedSTaS} and its ability to achieve $\epsilon$-LDP. Our experiment results show that \textit{FedSTaS} is able to achieve accuracy scores higher than those of \textit{FedSTS} within a fixed number of rounds on both IID and non-IID data. 

\section{Methodology}
\subsection{Model Aggregation}
As mentioned, the goal of FL is to jointly learn a global model. Mathematically, if there are $N$ clients, the goal of FL is to solve the following optimization problem:
\begin{equation}
    \label{eq: FL goal}
    \begin{split}
        \min_{\bm w\in \mathbb{R}^d} F(\bm w) &= \min_{\bm w\in \mathbb{R}^d} \sum_{k=1}^{N}\omega_k F_k(\bm w) \\
        &= \min_{\bm w\in \mathbb{R}^d} \sum_{k=1}^{N}\omega_k \frac{1}{n_k}\sum_{j=1}^{n_k}f(\bm w;x_{k,j})\\
    \end{split}
\end{equation}
where $\omega_k$ is a weight for the $k$th client such that $\omega_k\ge0$ and $\sum_{k=1}^{N}\omega_k=1$, and $n_k$ is the size of the $k$th client's training data, $\mathcal{D}_k=\{x_{k,1},\dots,x_{k,n_k}\}$. Each client's objective function $F_k(\bm w)$ is simply an average of the loss function, $f(\bm w; x_{k,j})$, over the client's training data, using the model parameters $\bm w\in\mathbb{R}^d$.

During the $t$th round of training, the server sends $\bm w_t$, the global parameter, to each of the clients which then perform $E$ epochs of local training. For the $k$th client, this update is given by
\begin{equation}
    \label{eq: kth client local training}
    \bm w_{t+1}^k = \bm w_{t}^k - \eta_t\nabla F_k(\bm w_{t}, \xi_{t}^k),
\end{equation}
where $\eta_t$ is the learning rate, and $\xi_t^k$ is a sample of the $k$th clients training data. Then, each client communicates their updated $\bm w_{t+E}^k$ to the server which are aggregated to compute the updated parameter. If each of the $N$ clients participate in round $t$, the full aggregation is computed as
\begin{equation}
    \label{eq: full participation aggregation}
    \bm w_{t+E} = \sum_{k=1}^{N}\omega_k\bm w_{t+E}^k.
\end{equation}

This global aggregation is naturally computationally inefficient and thus in each round, a random subset of $m$ clients, $S_t$, is chosen to be used in the aggregation. That is,
\begin{equation}
    \label{eq: sample participation aggregation}
    \bm w_{t+E} = \sum_{k\in S_t}\frac{N}{m}\omega_k w_{t+E}^k.
\end{equation}

In FL literature, a client sampling scheme is referred to as \textit{unbiased} if the following result holds:
\begin{equation}
    \label{eq: unbiased sampling}
    \mathbb{E}\left[\sum_{k\in S_t}\frac{N}{m}\omega_k w_{t+E}^k\right] = \sum_{k=1}^{N}\omega_k\bm w_{t+E}^k,
\end{equation}
where the expectation is taken with respect to the client sampling $S_t$.

\subsection{Client Level Sampling}
In \textit{FedSTS} \cite{FedSTS}, the authors propose to stratify the $N$ clients into $H$ non-overlapping strata based on the similarity of the Information Squeezed (IS) gradients of each client. Afterwards, they apply a sample size re-allocation scheme to optimally choose the number of clients to sample from each strata, $m_h$. Lastly, they sample each client with probabilities proportional to the norm of each client's gradient. This results in clients with a higher norm of gradient being more likely to be sampled, "assigning more attention to the more representative clients" as stated in \textit{FedSTS} \cite{FedSTS}.

To illustrate this method, let $\bm w_{h_i}$ denote the model update of the $i$th client in the $h$th stratum, let $N_h$ be the number of clients in the $h$th stratum, and let $\bm w_h=\frac{1}{m_h}\sum_{i=1}^{m_h}\bm w_{h_i}$ be the sampled average model update for the $h$th stratum. Then, the \textit{FedSTS} sampled model aggregation is given by
\begin{equation}
    \label{eq: FedSTS sample aggregation}
    \bm w_{sts} = \frac{1}{N}\sum_{h=1}^{H}N_h\bm w_h.
\end{equation}

Pseudo code for the client stratification proposed in \cite{FedSTS} is given in Algorithm \ref{alg: client stratification}.

\begin{algorithm}
    \SetAlgoLined
    \caption{ClientStratification}
    \label{alg: client stratification}
    \KwIn{Compressed gradient of all clients in the $t$th round $\{X_k^t\}_{k=1}^N$ and corresponding cluster index $\{\lambda_k^t\}_{k=1}^N$}
    \KwIn{The number of strata $H$}
    \BlankLine
    \textbf{Initialize} Use $\lambda_k^t$ to restore $\{X_k^t\}_{k=1}^N$ as $\{Z_k^t\}_{k=1}^N$\;
    
    \textbf{Initialize} Randomly select $H$ clients as group or stratum centers $\{\mu_1, \mu_2, \ldots, \mu_H\}$\;
    
    \textbf{Initialize} $C_i = \emptyset \; (1 \leq i \leq H)$\;
    
    \Repeat{$\forall\ i=\{1,2,\dots,H\},\mu_i'=\mu_i$}{
        \ForEach{client $k \leq N$}{
            $\epsilon_k = \arg \min_{i=1,2,\ldots,H} \|Z_k^t - \mu_i\|_2$\;
            
            $C_{\epsilon_k} = C_{\epsilon_k} \cup \{k^{th}\text{ client with } Z_k^t\}$\;
        }
        \ForEach{stratum $C_i,\ i \leq H$}{
            New center $\mu_i' = \frac{1}{|C_i|} \sum_{Z_k^t \in C_i} Z_k^t$\;
            
            $\mu_i \gets \mu_i'$\;
        }
    }
    \KwOut{$\mathcal{G} = \{C_1, C_2, \ldots, C_H\}$}
\end{algorithm}

\subsection{Data Level Sampling}
In \textit{FedSampling} \cite{FedSampling}, the server opts to sample data uniformly across the clients, resulting in a sampling scheme that more closely aligns with a centralized learning environment. To accomplish this, the data are sampled proportional to the required sample size and the total number of samples across all clients. Usually the client data size is a sensitive attribute, and thus privacy preservation is of importance. In \textit{FedSampling} \cite{FedSampling} the authors provide a privacy-preserving ration estimation method for estimating the total sample size, while ensuring the privacy of the client's data size. 

\subsection{FedSTaS: Federated Stratification and Sampling}
In this section we present the workflow for our proposed \textit{FedSTaS} federated training method who's pseudo code is shown in Algorithm \ref{alg: FedSTaS}. 

From a client sampling standpoint, we employ \textit{FedSTS} \cite{FedSTS}. That is, we stratify the clients based on their compressed gradient information, compute optimal re-allocations based on the size and gradient variance of the strata, and sample clients using importance sampling based on the norm of their gradients. Then, following the client sampling, we propose to conduct a second round of data-level sampling from the participating clients.

\begin{algorithm}
    \SetAlgoLined
    \caption{FedSTaS}
    \label{alg: FedSTaS}
    \KwIn{Updates in the $t$th round of all clients $\{G_t^k\}_{k=1}^{N}$}
    \KwIn{Desired client sample size $m$ and desired data sample size $\eta$}
    \BlankLine
    \textbf{Initialize} $\bm w_0$\;
    
    \textbf{Initialize} $S_t=\emptyset$\;

    \ForEach{round $t=1,2,\dots$}{
        $\mathcal{G} = $ ClientStratification$(\bm X_t, \lambda_t, H)$\;
        
        \ForEach{stratum $h\le H$}{
            $m_h = m\cdot \frac{N_hS_h}{\sum_{h=1}^{H}N_hS_h}$\;
            
            $p_t^k = \frac{||IS(G_t^k)||}{\sum_{k=1}^{N_h}||IS(G_t^k)||}, k=1,\dots,N_h$\;
            
            $S_t=S_t\cup m_h$ clients sampled with $\{p_t^k\}_{k=1}^{N_h}$\;
        }
        \ForEach{sampled client $i\in S_t$}{
            compute $r_{h_i}$ using \ref{eq: dp client size}\;
        }
        compute $\tilde{n}$ using \ref{eq: dp data size}\;
        
        \ForEach{sampled client $k\in S_t$}{
            $\bm w_{t+1}^k$ = ClientUpdate$(k,\bm w_t, \eta/\tilde{n})$\;
        }
        $\bm w_{t+1}= \frac{1}{N}\sum_{h=1}^{H}N_h\frac{1}{m_h}\sum_{k=1}^{m_h}\bm w_{t+1}^k$
    }
\end{algorithm}

Our data-level sampling strongly resembles that of the \textit{FedSampling} \cite{FedSampling}, but is adjusted to mimic centralized learning across the participating clients rather than the whole set of clients. Suppose that in the $t$th round, the server wishes to sample $\eta$ observations across the $m$ sampled clients. With $n_{h_i}$ denoting the size of the training data for the $i$th client in the $h$th stratum, let $n=\sum_{h=1}^{H}\sum_{i=1}^{m_h}n_{h_i}$ denote the total number of observations in the set of participating clients. Then, without preserving privacy, we propose to sample the data uniformly across the participating clients with sampling probabilities equal to $\eta/n$.

To preserve privacy, however, we follow \textit{FedSampling} \cite{FedSampling}. First, each participating client will compute a clipped sample size, $n_{h_i,c}=\min(n_{h_i}, M-1)$ for some size threshold $M$. Following this, a fake sample size is drawn from a uniform Multinomial distribution, $\hat{n}_{h_i}\sim \mathcal{P}(M)$. Next, $x_{h_i}$ is drawn from a Bernoulli distribution with parameter $\alpha$ to generate the following privacy preserving client data size:
\begin{equation}
    \label{eq: dp client size}
    r_{h_i} = x_{h_i}n_{h_i,c}+(1-x_{h_i})\hat{n}_{h_i}.
\end{equation}

Lastly, we compute a private version of the total participating data size $n$:
\begin{equation}
    \label{eq: dp data size}
    \tilde{n}=\left(R-\frac{(1-\alpha)Mm}{2}\right)/\alpha,
\end{equation}
where $R=\sum_{h=1}^{H}\sum_{i=1}^{m_h}r_{h_i}$. $\tilde{n}$ is then used to sample the data uniformly across the participating clients with sampling probabilities equal to $\eta/\tilde{n}$. This data level sampling scheme is illustrated in Algorithm \ref{alg: client update}.

\begin{algorithm}
    \SetAlgoLined
    \caption{ClientUpdate}
    \label{alg: client update}
    \KwIn{Client $k$}
    \KwIn{Model parameter $\bm w_t$}
    \KwIn{Sampling ratio $\eta/\tilde{n}$}
    \BlankLine
    
    \textbf{Initialize} $\bm w_{t}^k = \bm w_t$\;
    
    $\xi_{t}^k\leftarrow$ a sample of $(\eta/m)$ observations from $\mathcal{D}_k$ taken with sampling probabilities $p_{k_j}=\eta/\tilde{n}$, $j=1,\dots,n_k$\;
    
    \ForEach{epoch $i=1,\dots,E$}{
        $\bm w_{t+i+1}^k = \bm w_{t+i}^k-\eta\nabla F_k(\bm w_{t+i}^k,\xi_{t}^k)$\;
    }
    \KwOut{$\bm w_{t+1}^k = \bm w_{t+E}^k,X_t,\lambda_t=IS(G_t^k)$}
\end{algorithm}

\subsection{Theoretical Results}
In this subsection we reference the theoretical properties of our proposed \textit{FedSTaS} method. Firstly, from a client sampling perspective, our method is unbiased. This is an immediate consequence of the same result in \textit{FedSTS} \cite{FedSTS}.

\begin{lemma}[Unbiased-ness]
    Let $\bm w_{t+1}$ be computed via Algorithm \ref{alg: FedSTaS}. Then 
    \begin{equation*}
        \mathbb{E}_{S_t}[\bm w_{t+1}] = W(\mathcal{K}),
    \end{equation*}
    where $W(\mathcal{K})$ denotes the model aggregation computed with all clients.
\end{lemma}

This lemma indicates that our method does not introduce bias into the global model, ensuring that our sample aggregation is representative of the total population. Furthermore, from a slight modification of Lemma 3.1 from \cite{FedSampling} we have the following lemma.

\begin{lemma}
    Let $p = \eta/n$ and $\tilde{p}=\eta/\tilde{n}$ denote the data level sampling probabilities for centralized learning on the sampled clients, and the proposed method, respectively. Then, as the number of sampled clients increases, the mean square error between $p$ and $\tilde{p}$ converges to 0. That is, 
    \begin{equation}
        \label{eq: asymp. mse}
        \lim_{m\to\infty}\mathbb{E}[(\tilde{p} - p)^2]=0.
    \end{equation}
\end{lemma}
The proof of this lemma is directly analogous to that of \cite{FedSampling} with the slight adjustment of focusing on the participating clients instead of the set of all clients. The lemma provides the guarantee that as the set of participating clients increases, our training method more closely mimics that of centralized learning. 

Our final theoretical result is also derived via a slight modification of Lemma 3.2 from \cite{FedSampling}. This lemma is relevant to the privacy protection of the client's local sample sizes. The lemma states that $\tilde{n}$ computed from \ref{eq: dp client size} can achieve $\epsilon$-LDP.

\begin{definition}[$\epsilon$-LDP]
    A random mechanism $\mathcal{M}$ satisfies $\epsilon$-LDP if and only if for two arbitrary inputs $x$ and $x'$, and any output $y$ in the image of $\mathcal{M}$,
    \begin{equation}
        \frac{\mathbb{P}(\mathcal{M}(x)=y)}{\mathbb{P}(\mathcal{M}(x')=y)} \le e^{\epsilon}.
    \end{equation}
\end{definition}

\begin{lemma}
    Given an arbitrary size threshold $M$, our privacy preserving estimation of the client's local sample sizes achieves $\epsilon$-LDP, when
    \begin{equation}
        \alpha = \frac{e^{\epsilon}-1}{e^{\epsilon}+M-2}.
    \end{equation}
\end{lemma}
\section{Experiments}\label{sec:experiments}

\subsection{Experimental Setup}

We conducted experiments under both IID and non-IID data distributions to evaluate the performance of our federated learning approaches. The dataset was partitioned into 100 clients. For the IID setting, we performed a uniform random split, ensuring each client received an approximately equal portion of the data. For the non-IID setting, we employed a Dirichlet distribution (\(\text{Dir}(\alpha)\)) to simulate heterogeneous label distributions across clients. The smaller the parameter \(\alpha\), the more skewed the resulting distribution. In our experiments, we set \(\alpha = 0.01\) to create a highly imbalanced and challenging scenario for federated learning.

\begin{figure}[ht]
    \centering
    \includegraphics[width=0.75\linewidth]{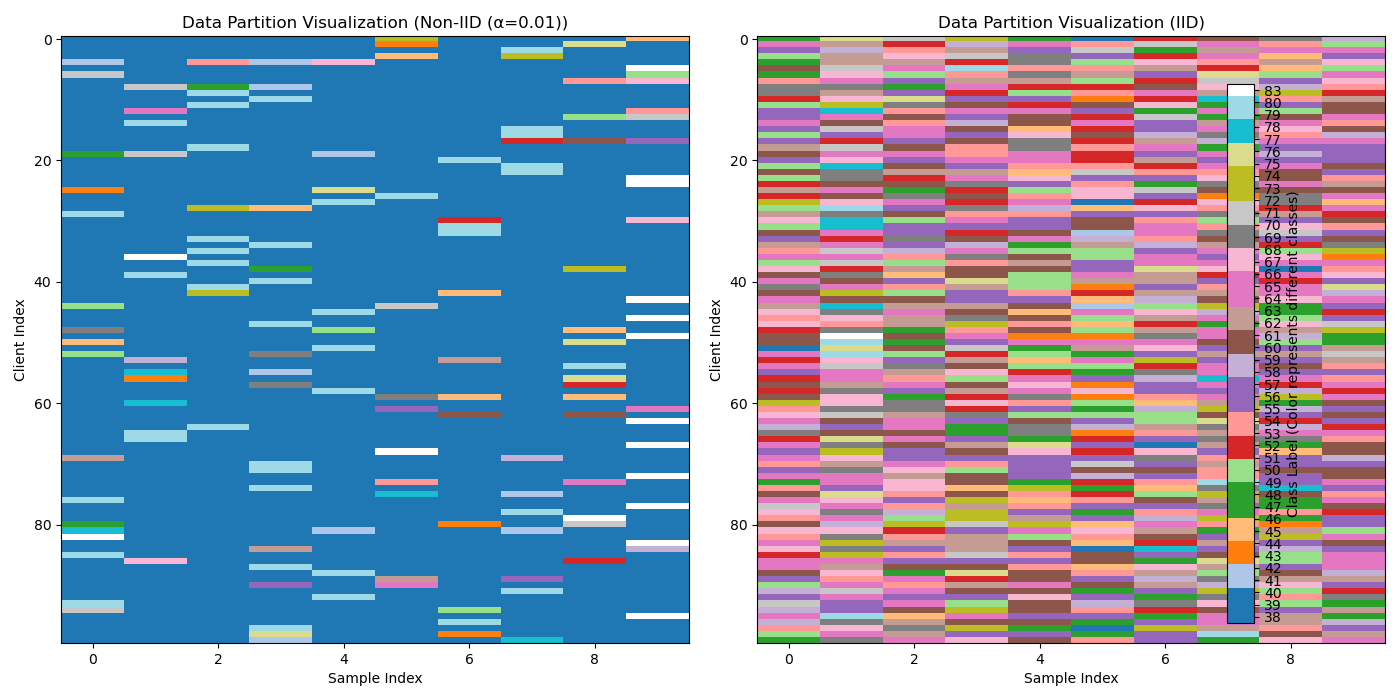}
    \caption{Visualization of data partition for Dirichlet distribution with \(\alpha = 0.01\) and IID settings.}
    \label{fig:data_partion}
\end{figure}

Figure~\ref{fig:data_partion} visualizes the data partition under both IID and non-IID distributions with \(\alpha = 0.01\). The non-IID case clearly exhibits a more uneven distribution of labels compared to the IID setting, closely mirroring realistic federated learning environments.

We evaluated our methods on two popular benchmark datasets: MNIST and CIFAR-100. MNIST comprises 70,000 grayscale handwritten digit images (60,000 training and 10,000 testing) distributed across 10 classes. CIFAR-100 consists of 60,000 color images (50,000 training and 10,000 testing) spread over 100 classes. For non-IID data, the choice of \(\alpha = 0.01\) created substantial imbalance among clients.

Our experimental setup considered a non-convex optimization setting. We set the number of strata \(H = 10\) and sampled \(m = 10\) clients per iteration. The sampling ratio was \(q = 0.1\), determining the requested sample size \(\eta\). Models were trained for \(T = 99\) and \(T = 199\) communication rounds. The key hyperparameters were as follows: \(q = 0.1\), \(n_{\text{SGD}} = 3\), \(\eta = 0.01\), \(B = 128\) for MNIST, \(\alpha = 0.01\), \(n_{\text{iter}} = 99\), \(K_{\text{desired}} = 2048\), \(d' = 9\), \(M = 100\), and \(\alpha_{\text{dp}} = 0.1616\) to achieve a differential privacy (DP) guarantee of \(\epsilon = 3\).


\subsection{Results and Analysis}

We investigated three key methods: \textbf{Stratified (FedSTS)}, \textbf{Compressed Gradients (FedSTaS)}, and \textbf{DP + Compressed Gradients (DP + FedSTaS)}. Our primary focus was on their convergence behavior and final accuracy under challenging non-IID conditions. Note that the Compressed Gradients approach is our proposed method, hereafter referred to as FedSTaS, which builds upon the baseline (FedSTS) by incorporating gradient compression techniques to enhance efficiency and performance.

\subsubsection{Performance on MNIST (\(\alpha = 0.01\))}

\begin{figure}[ht]
    \centering
    \begin{subfigure}[b]{0.49 \linewidth}
        \centering
        \includegraphics[width=\linewidth]{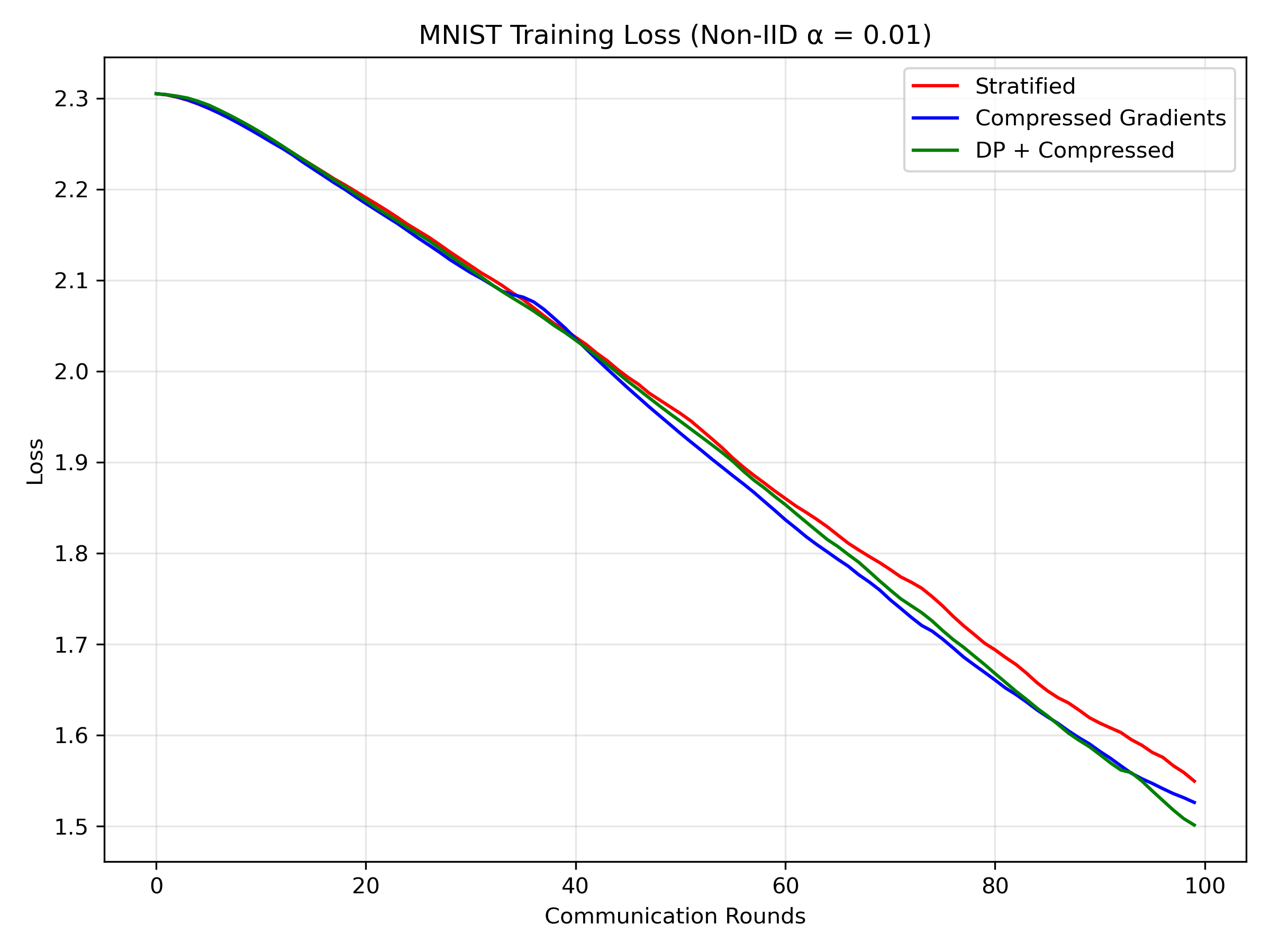}
        \caption{Training Loss}
    \end{subfigure}
    \begin{subfigure}[b]{0.49 \linewidth}
        \centering
        \includegraphics[width=\linewidth]{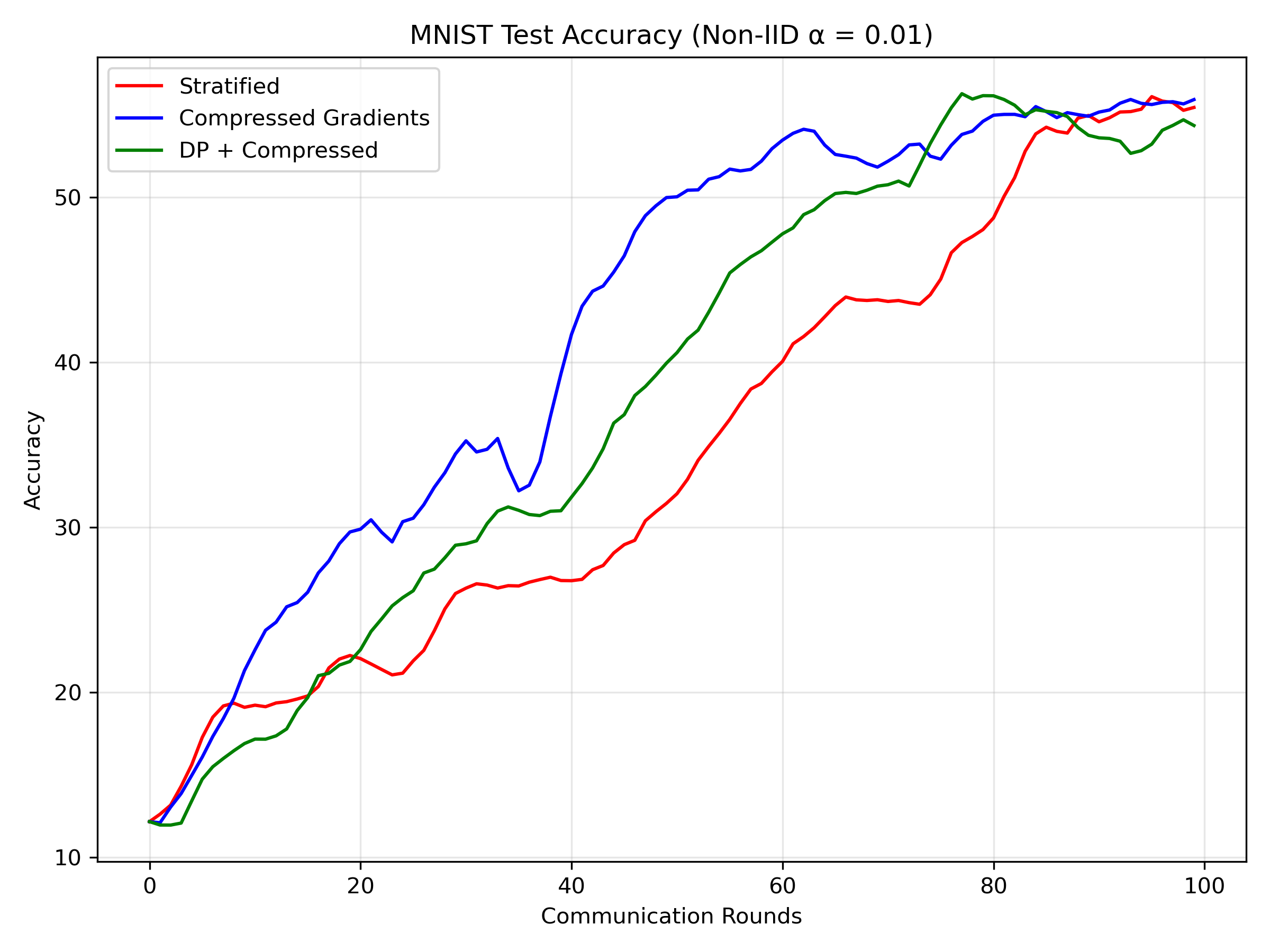}
        \caption{Test Accuracy}
    \end{subfigure}
    \caption{Non-Convex model with \( q = 0.1 \), \( n_{\text{SGD}} = 3 \), \( \eta = 0.01 \), \( B = 128 \) on MNIST, with \( \alpha = 0.01 \), \( n_{\text{iter}} = 99 \), \( K_{\text{desired}} = 2048 \), \( d' = 9 \), \( M = 100 \), and \( \alpha_{\text{dp}} = 0.1616 \) (DP Privacy = 3).}
    \label{fig:mnist_alpha001}
\end{figure}

Figure~\ref{fig:mnist_alpha001} illustrates the training loss and test accuracy on MNIST with \(\alpha = 0.01\) over 99 communication rounds. As all methods display decreasing training loss, their primary differences emerge in the accuracy curves, particularly in terms of convergence speed and final accuracy.

\textbf{Stratified (FedSTS)}: The baseline method begins with approximately 10\% test accuracy and progresses steadily to around 50\% after 100 iterations. Despite this improvement, the convergence speed and final performance are moderate, reflecting the challenges posed by the non-IID data distribution.

\textbf{FedSTaS (Compressed Gradients)}: Our proposed compressed gradient method exhibits a substantially faster rate of improvement. It surpasses the FedSTS baseline by round 40 and continues to gain, achieving about 56\% accuracy. This robust performance underlines the capacity of gradient compression to effectively handle heterogeneous data distributions, leading to both faster convergence and higher final accuracy.

\textbf{DP + FedSTaS (DP + Compressed Gradients)}: Incorporating differential privacy slightly reduces the final accuracy compared to FedSTaS alone, reaching roughly 54.7\%. However, this level of performance still surpasses FedSTS, demonstrating that meaningful privacy guarantees can be integrated without entirely sacrificing the benefits of gradient compression.

Overall, these results attest to the effectiveness of our proposed FedSTaS method in accelerating convergence and enhancing accuracy in federated learning under non-IID conditions, while also showing that adding privacy preservation (DP + FedSTaS) remains a viable and competitive option.


\subsubsection{Performance on MNIST (\(\alpha = 0.001\)) and CIFAR-100 (\(\alpha = 0.001\))}

\begin{figure}[ht]
    \centering
    \includegraphics[width=0.7\linewidth]{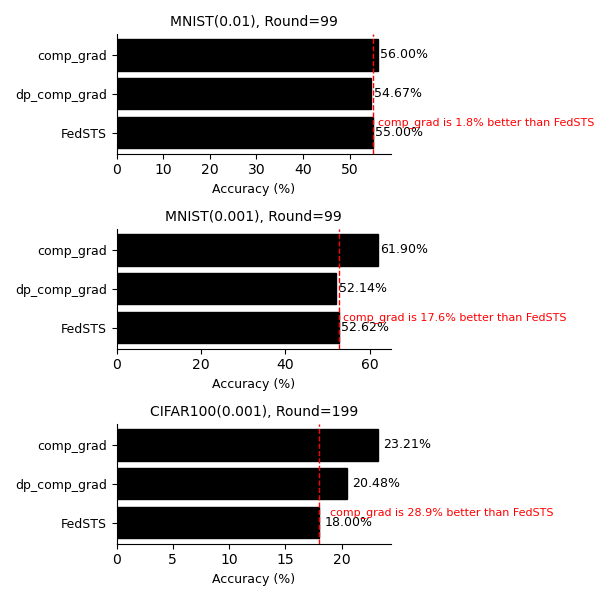}
    \caption{Non-Convex model with \( q = 0.1 \), \( n_{\text{SGD}} = 3 \), \( \eta = 0.01 \), \( B = 128 \) on MNIST and CIFAR-100 with varying \(\alpha\) values (0.01 and 0.001). The desired dimension is \( K_{\text{desired}} = 2048 \), \( d' = 9 \), \( M = 100 \), and \(\alpha_{\text{dp}} = 0.1616\) ensures DP Privacy \( \epsilon = 3 \).}
    \label{fig:mnist_cifar_all}
\end{figure}

To further investigate the robustness of the evaluated methods, we extended our experiments to more heterogeneous data settings (\(\alpha = 0.001\)) and to the more challenging CIFAR-100 dataset. Figure~\ref{fig:mnist_cifar_all} presents a comparative summary of the final test accuracies achieved by FedSTS (Stratified), FedSTaS (our Compressed Gradients approach), and DP + FedSTaS (Compressed Gradients with Differential Privacy) under these conditions. The figure comprises three horizontal bar charts: 
\begin{itemize}
    \item \textbf{Top}: MNIST with \(\alpha = 0.01\) at 99 rounds.
    \item \textbf{Middle}: MNIST with \(\alpha = 0.001\) at 99 rounds.
    \item \textbf{Bottom}: CIFAR-100 with \(\alpha = 0.001\) at 199 rounds.
\end{itemize}

For MNIST with \(\alpha = 0.01\) after 99 rounds (top chart), FedSTaS (our Compressed Gradients method) achieves 56.00\%, improving upon FedSTS by 1.8\%, while DP + FedSTaS reaches 54.67\%, still outperforming FedSTS.

When we decrease \(\alpha\) to 0.001 for MNIST (middle chart), the non-IID nature of the data intensifies. Under these more heterogeneous conditions, FedSTaS further amplifies its advantage, achieving 61.90\% accuracy after 99 rounds, surpassing FedSTS by a remarkable 17.6\%. Even with the added privacy constraints, DP + FedSTaS attains 52.14\%, maintaining an edge over FedSTS.

Turning to CIFAR-100 (bottom chart), with \(\alpha = 0.001\) after 99 rounds, we observe that although the absolute accuracies are lower due to the complexity of this 100-class classification task, the relative improvements remain notable. FedSTaS reaches 23.21\%, outperforming FedSTS by 28.9\%. Similarly, DP + FedSTaS achieves 20.48\%, still surpassing FedSTS by a significant margin despite the complexity and privacy constraints.

These results highlight that FedSTaS (our Compressed Gradients method) consistently delivers the fastest convergence and highest accuracies across varying levels of data heterogeneity and complexity. The DP + FedSTaS variant demonstrates that privacy-preserving measures can be integrated without entirely negating the benefits of gradient compression.


\section{Conclusion and Future Work}
Our experimental results confirm that Compressed Gradients achieve both faster convergence and higher accuracy compared to Stratified methods in non-convex federated learning scenarios, especially under non-IID conditions. Incorporating differential privacy (DP + Compressed Gradients) retains most of these benefits while ensuring a privacy guarantee of \(\epsilon = 3\). These findings highlight the practicality of gradient compression techniques for efficient and secure model training across diverse federated learning environments.

The first direction for future work is to compare the variance of the proposed method to other aggregation methods such as those proposed in \cite{FedSTS}, \cite{FedClust}, or \cite{FedStrat}, for example. Stemming from this, theoretical results on the convergence rate of our method are desirable. Further analysis on the privacy protecting properties of our method would be interesting and important for practical implementations.

Moreover, there exists possible improvements upon our method, such as optimal allocation for the data level sampling. This type of change could improve the variance of the data level sampling, possibly leading to better overall convergence rate.


\bibliography{references}
\bibliographystyle{icml2025}

\end{document}